\documentclass[a4paper]{article}
\usepackage{color, colortbl}
\usepackage{ amssymb }
\usepackage{xcolor}
\newcommand{\mn}[1]{{\textcolor{black}{#1}}{\bf}}
\newcommand{\pp}[1]{{\textcolor{black}{#1}}{\bf}}
\newcommand{\mg}[1]{{\textcolor{black}{#1}}{\bf}}

\newcommand{\mnn}[1]{{\textcolor{black}{#1}}{\bf}}

\usepackage{soul}
\newcommand{\COMMENT}[1]{}

\usepackage{INTERSPEECH2020}

\title{Contextualized
Translation of Automatically Segmented Speech}
\name{Marco Gaido$^{1,2}$, Mattia A. Di Gangi$^{1,2}$, Matteo Negri$^1$, Mauro Cettolo$^1$, Marco Turchi$^1$}
\address{
  $^1$Fondazione Bruno Kessler, Italy\\
  $^2$University of Trento, Italy}
\email{\{mgaido,digangi,negri,cettolo,turchi\}@fbk.eu}

\begin{document}

\maketitle
\begin{abstract}
Direct speech-to-text translation (ST) models are usually trained on corpora segmented at sentence level,
but at inference time they are commonly fed with audio split by a voice activity detector (VAD).
Since VAD segmentation is not syntax-informed,
the resulting segments do not necessarily correspond to well-formed sentences uttered by the speaker
but, most likely, to fragments of one or more sentences.
This segmentation mismatch degrades considerably the quality of ST models' output. 
So far, researchers have focused on improving 
audio segmentation
towards producing sentence-like splits.
In this paper, instead, we address the issue  in the model, making it
more robust to a different, potentially  sub-optimal segmentation.
To this aim, we train our models on randomly segmented data
and 
compare two approaches: fine-tuning and adding the previous segment as context.
We show that our 
context-aware solution is more robust to VAD-segmented input, outperforming
a strong base model and the fine-tuning on different
VAD segmentations of an English-German test set by up to 4.25 BLEU points.

\end{abstract}


\section{Introduction}
\label{sec:intro}
Speech-to-text translation (ST) has been traditionally addressed by pipeline approaches
involving several components \cite{iwsltprec_2019}.
The most important blocks are the speech recognition (ASR),
which converts the input audio into its transcript,
and the neural machine translation (NMT),
which translates the transcript into the target language.
Direct ST models \cite{berard_2016,weiss2017sequence} recently gained attention as an alternative approach, thanks to their appealing promises to overcome some of the pipeline systems' problems, such as error propagation and loss of information present in the audio (prosody in particular).

Both pipeline and direct solutions, however, can be significantly affected by mismatches in the segmentation of the input between training and test data.
On one side, the two solutions involve the use of training data segmented at sentence level. This, for instance, holds for the parallel corpora normally used to train the NMT component of the pipeline approach, as well as for all the available ST corpora used for 
direct ST
training.
On the other side, at inference time both solutions will be exposed to data segmented according to criteria that look at properties of the audio input rather than at linguistic notions like sentence well-formedness. The most widespread approach consists in fact in using voice activity detection (VAD) to split the audio stream into chunks,
which are input to the ST system.
In particular, VAD systems determine
whether a given short (usually 10-30 ms) audio segment 
actually contains speech,
and this information is used in the context of ST for two purposes:
\textit{i)} dividing the audio stream into segments containing uninterrupted speech;
\textit{ii)} filtering out audio
segments containing other sounds.

Since VAD is solely based on the alternation between human voice, silences 
and other sounds, the resulting splits might not correspond to well-formed sentences but to fragments of one or more sentences.
The impact of feeding an ST model trained on ``clean'' data with sub-optimal, not linguistically-motivated segmentations varies according to the characteristics of the VAD employed and its settings. Very aggressive settings reduce the generation of long (cross-sentential) segments,
which are difficult to handle by neural models that are typically very sensitive to input length.
On the downside, they produce 
short (sub-sentential) segments that might not provide enough context for 
proper translation.
To address this problem, pipeline systems include an additional component that
re-segments the ASR output to 
provide the NMT 
with well-formed sentences \cite{matusov_segm, oda-etal-2014-optimizing, Cho2017}.
Since this solution is not possible for direct 
ST, 
where
the two steps are not decoupled, 
researchers have worked on 
alternative audio segmentation techniques.
In the 2019 IWSLT offline ST task \cite{iwsltprec_2019}, for instance,
the best direct ST system \cite{potapczyk_tomasz_2019} had one of its key features in the 
segmentation method.

Instead of working on the segmentation algorithm, in this paper we aim to make our direct ST models more robust to VAD-segmented data.
To  train them on a data distribution more similar to the one fed at inference time, we generate an artificial dataset by  
randomly 
re-segmenting clean (i.e. sentence-based) ST data.
Then, we experiment with two approaches: \textit{i)} 
fine-tuning on
the new dataset;
\textit{ii)} 
improving our direct ST model with the capability to look back and attend to the preceding segment as contextual information.
Our experiments 
show that the proposed context-based solution 
effectively handles the segmentation of different VAD systems and configurations,
reducing the drop in translation quality caused by
segmentation mismatches in the training and test data by up to 55\%.

\section{Context-aware ST}

The idea of exploiting contextual information to improve translation 
has been successfully applied in NMT
\pp{\cite{wang-etal-2017-exploiting-cross,zhang-etal-2018-improving,bawden-etal-2018-evaluating,kim-etal-2019-document}}.
In our use case, unlike \cite{wang-etal-2017-exploiting-cross},
we are interested only in modeling short-range cross-segment dependencies
\mn{to cope with the sub-optimal}
breaks introduced by 
VAD
segmentation.
We hence consider as context only the segment
immediately preceding the one to be translated, leaving
out of our study hierarchical 
approaches modeling the whole document  as context.
Moreover, while in document-level NMT the best approach is to use the source side of the sentence(s) as contextual information,
in the ST scenario it is not trivial to understand which side is best.
\pp{On one hand,}
audio source avoids the error propagation and exposure bias introduced by using as context the translations generated at inference time.
\pp{On the other,} these problems are balanced by the easiness of extracting information from text rather than from audio \cite{instance_based}. 
In this work,  we study both options.

To integrate context information into the model, we explore the two solutions that gave the best results for NMT \cite{kim-etal-2019-document}.
\mg{They} respectively use  sequential \cite{zhang-etal-2018-improving} and  parallel \cite{bawden-etal-2018-evaluating} decoders.
We also experimented with the integration of 
context information in the encoder \cite{zhang-etal-2018-improving}, but
the trainings were either very unstable (when using audio as context)
or ineffective, \pp{eventually} leading to worse results. For this reason, we do not consider
this type of integration in the rest of the paper.
Finally, supported by previous findings \cite{kim-etal-2019-document},
we 
\pp{neither}
investigate the concatenation of the context with the current input \cite{agrawal_ctx}, \pp{nor}
the combination of encoded representations of 
\mn{the two}
\cite{voita-etal-2018-context}.

\pp{Our base model is an adaptation of Transformer \cite{transformer}:
\mg{its} encoder is enhanced
to take into account the characteristics of speech input 
by means of two 2D convolutional layers and a logarithmic distance penalty in its self-attention 
layers \mnn{\cite{digangi:interspeech19}}.}
Both the sequential and the parallel decoder use a multi-encoder approach,
with an additional encoder dedicated to the context information.
\mg{However,} they differ in the way this information is integrated into the base model.
The context encoder is composed of Transformer encoder layers,
but its input depends on the modality of the segment used as context, i.e. text or audio.
When we use the generated translations as context, 
 its tokens are converted into vectors with
 \textit{word embeddings} (namely, we re-use the decoder embeddings), summed with  \textit{positional encoding} and then provided to the encoder Transformer layers.
When we use the audio as context, the input audio features are first processed by the encoder of the base model and then 
passed to the context encoder \cite{instance_based}.

\begin{figure}[htbp]
\centering
\includegraphics[width=0.25\textwidth]{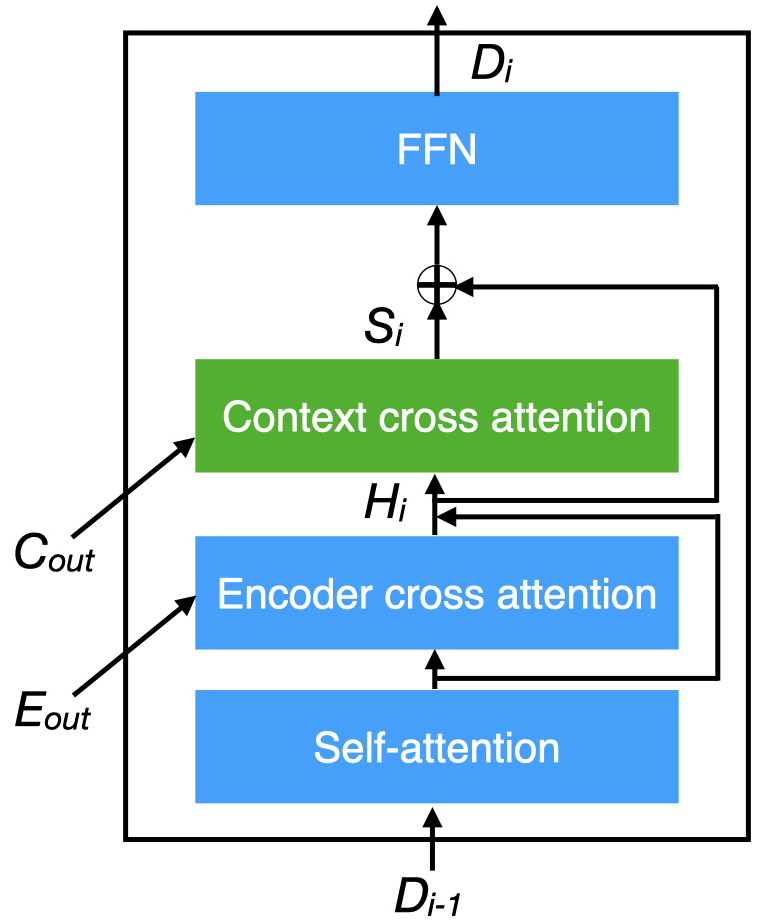}
\caption{Sequential context integration.}
\label{fig:seq_arch}
\end{figure}

\noindent \textbf{Sequential} (Figure \ref{fig:seq_arch}). In each decoder Transformer layer,
an additional multi-head cross-attention sub-layer is introduced.
It queries the output $C_{out}$ of the context encoder
using the output $H_i$ of the $i$-th encoder cross-attention sub-layer. 
The result $S_i$ of this operation is combined with $H_i$ using a position-wise gating mechanism, before being fed to the feed-forward network $\text{FFN}_i$.
Hence, the output of the $i$-th decoder layer $D_i$ is:
\begin{equation}
  \lambda_i = \sigma (W_{hi} H_i + W_{si} S_i)
  \label{gating_lambgda_eq}
\end{equation}
\begin{equation}
  D_i = \text{FFN}_i (\lambda_i H_i + (1 - \lambda_i) S_i)  
  \label{decoder_gating_eq}
\end{equation}

\begin{figure}[htbp]
\centering
\includegraphics[width=0.45\textwidth]{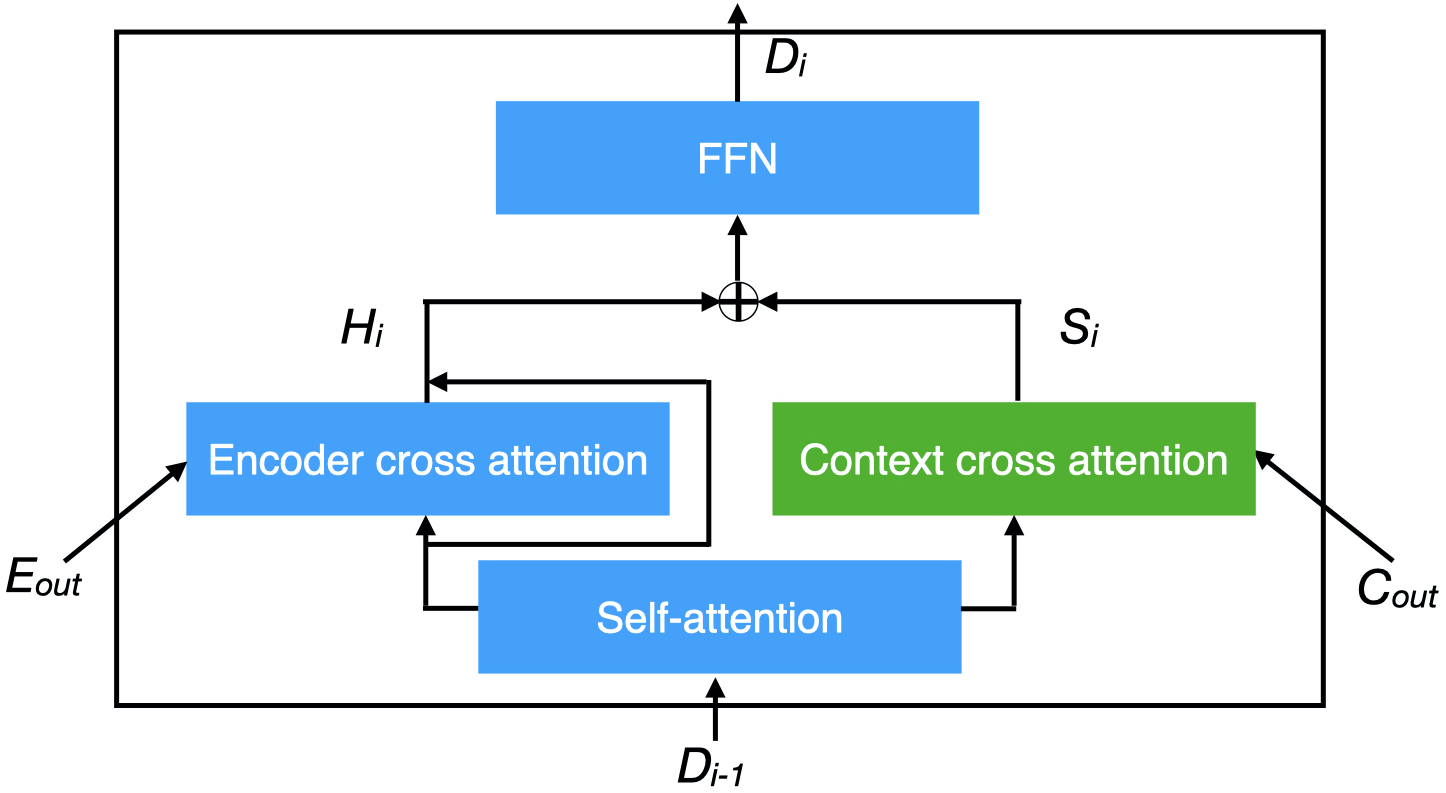}
\caption{Parallel context integration.}
\label{fig:par_arch}
\end{figure}

\noindent \textbf{Parallel} (Figure \ref{fig:par_arch}). In each decoder Transformer layer, the output of the self-attention sub-layer is used as
query for both the encoder cross-attention and the context cross-attention defined in the same way as in the previous case.
The outputs of these two sub-layers are then combined using the
position-wise gating mechanism described in Eq.(\ref{decoder_gating_eq}).

To 
avoid over-relying on
the context,
we add a regularization on the context gate. 
Our regularization is slightly different from the one proposed by \cite{li2019regularized}:
we always penalize the context information,
so that the model will
use it only when it is strictly needed.
With the regularization factor, the resulting loss is:

\begin{equation}
  \mathcal{L'} = \mathcal{L} + \alpha \sum_{i=0}^{N_d}(1 - \lambda_i)
  \label{regularization_gate_eq}
\end{equation}

\section{\pp{Experimental settings}}

\subsection{\pp{Clean and artificial data}}

\pp{Our base models are trained and evaluated on English-German data drawn from the MuST-C corpus, the largest ST dataset currently available \cite{mustc}. MuST-C comprises  234K samples (corresponding to about 408 hours of speech) divided into training (229K), validation (1.5K) and test sets (3.5K).}

\label{sec_data}
\pp{To cope with segmentation mismatches between the clean data used for training and the VAD-processed ones handled at inference time, we generate an automatic re-segmentation of the MuST-C training and validation set.}
The re-segmentation starts by picking a random (with uniform distribution) \textit{split word} for each sample in the original English 
\pp{transcripts.} Each fragment spanning from a \textit{split word} to the word before the next \textit{split word} becomes a segment of the new training set and the preceding fragment becomes its context. We extract the audio corresponding to each resulting transcript by leveraging word alignments computed with Gentle.\footnote{https://github.com/lowerquality/gentle/} Then, we retrieve the corresponding translations using word alignments generated with  fast\_align  \cite{dyer-etal-2013-simple}. In case of missing alignments (either with the audio or with the translation), the sample is discarded. The resulting training dataset contains 225K samples (4K less than the original), while 
the validation set size  is almost unchanged.

A manual check on a sample of the produced  aligned segments revealed that about 96\% of  them are acceptable. The  most frequently observed issue  is that some translations  contain 1-2 words more than the optimum, mostly due to the lack of some word alignments and to word-reordering. This  leads to the presence of overlapping words between the context and the target German references in 25\% of the samples. In early experiments, this caused model instability at inference time because models learnt to copy the final context words, up to producing nonsensical sequences of repeated tokens. We solved the issue by filtering out the overlapping words from the context.

\subsection{VAD and segmentation}
As we want our systems to be robust to different VAD 
outputs, we test our models
on two different open source VAD tools:
LIUM \cite{meigner2010lium}
and WebRTC's VAD.\footnote{https://webrtc.org/. We use the open-source Python interface https://github.com/wiseman/py-webrtcvad}
For WebRTC we tested all the possible configurations, varying the \textit{frame size}
(allowed values are $10$ms, $20$ms and $30$ms) and the \textit{aggressiveness}
(ranging from $0$ to $3$, extremes included).
\mg{We discarded those}
producing either too long ($>60$s) or too many segments
($>5,100$, i.e. twice the segments of the original sentence-based segmentation \pp{of the MuST-C test set}).
In this way,
 we ended up with three configurations,
whose characteristics are described in Table \ref{tab:vad_summary}.

Overall, the segments produced by WebRTC have much higher variance in their length (ranging from $0.40$s to $58.62$s) compared to LIUM (from $2.50$s to $18.63$s) and are significantly more ($>3,500$ vs $2,725$).
As anticipated in $\S\ref{sec:intro}$, this can affect the final performance of neural ST models, for which handling very long/short segments is
difficult.
However, from a qualitative standpoint, 
a manual inspection of 50 samples showed that the split times selected by LIUM are less accurate than those selected by WebRTC: while the former 
often splits fluent speech, the latter always selects 
\mn{positions in which}
the speaker is silent. 

\begin{table}[htbp]
    \caption{Statistics for  different segmentations of the MuST-C test set. ``Man.'' refers to the original sentence-based segmentation.}
    \label{tab:vad_summary}
    \centering
    \begin{tabular}{l|c|c|ccc}
    \toprule
    \textbf{System} & Man. & LIUM & \multicolumn{3}{c}{WebRTC} \\
    \textbf{Frame size} & ~~~ & ~~~ & 30ms & 20ms & 20ms  \\
    \textbf{Aggress.} & ~~~ & ~~~ & 3 & 2 & 3  \\
    \midrule
        \textbf{\% filt. audio}  & 14.66 & 0.00 & 11.27 & 9.53 & 15.58 \\
        \textbf{Num. segm.}  & 2,574 & 2,725 & 3,714 & 3,506 & 5,005 \\
        \textbf{Max len. (s)}  & 51.97 & 18.63 & 48.84 & 58.62 & 46.76 \\
        \textbf{Min len. (s)}  & 0.05 & 2.50 & 0.60 & 0.40 & 0.40 \\        
    \bottomrule
    \end{tabular}
\end{table}

\subsection{Training settings}
\pp{All our models are optimized}
with label smoothed cross entropy \cite{szegedy2016rethinking}
using the Adam optimizer \cite{adam} with a learning rate starting
from 
$3\cdot10^{-4}$,
increasing linearly up to $5\cdot10^{-4}$  in the first $5,000$ steps and then decaying
with inverse square root policy.
The overall batch size was $512$ \textit{(audio, translation)} pairs.
We used the \textit{BIG} configuration from \mg{\cite{digangi:interspeech19}}
regarding all layers' hidden sizes. The number of context encoder layers $N_c$ is set to $1$, as \cite{zhang-etal-2018-improving} shows that this leads to the best results. Since~\cite{kim-etal-2019-document} has demonstrated that poorly regularized systems can
lead to ambiguous results when 
\mn{integrating context}, we used $0.2$ dropout
and \textit{SpecAugment} \cite{Park_2019} to prevent this issue.

We performed preliminary experiments on a baseline model 
(\textit{BASE\_MUSTC})
with $8$ encoder layers $N_e$ and $6$ decoder layers $N_d$ trained on the MuST-C En-De training set.
Since models using the generated translations as context are affected 
by exposure bias,
we wanted to test our solution also in more realistic conditions,
with a stronger baseline model trained in rich data conditions.
This model 
(\textit{BASE\_ALL})
was trained with $N_e$ set to $11$ and $N_d$ to $4$, on all the data available for the IWSLT 2020 evaluation campaign,\footnote{http://iwslt.org/doku.php?id=offline\_speech\_translation}
with knowledge distillation from an MT model and synthetic data generated translating the transcripts of ASR corpora.
Its training involves a pre-training on the synthetic data,
a fine-tuning on the data having ground-truth translations and
a second fine-tuning using label-smoothed cross entropy instead of knowledge distillation \mg{\cite{gaido-etal-2020-end}}.

All the context-aware models are initialized with the corresponding baseline model trained on sentence-segmented data.
We experimented with freezing all the pre-trained parameters as in~\cite{zhang-etal-2018-improving},
but freezing the decoder weights turned out to be harmful.
If freezed, decoder's layers are not able to adapt to the new inputs \mn{(with different segmentation)} and this slows down convergence
and leads to worse results.
\mg{We hence freeze  only the encoder.}
\mg{Our code is based on fairseq \cite{ott-etal-2019-fairseq} and is available at https://github.com/mgaido91/FBK-fairseq-ST.}

\mn{Textual data were}
pre-processed with tokenization and punctuation normalization
\mn{performed using}
Moses \cite{koehn-etal-2007-moses}, and 
\mn{were}
segmented with $8,000$ BPE merge rules \cite{sennrich2015neural}.
\mn{For the audio,}
we applied $40$ Mel filters
with window size of $25$ms and stride of $10$ms, \mn{performing} speaker normalization with XNMT~\cite{neubig-etal-2018-xnmt}.
To avoid out-of-memory errors, we excluded from the training set the audio segments longer than $20$ seconds.

\mn{In all cases, evaluation is performed on the best model according to the loss on the validation set.}
\mn{The  metrics used are BLEU \cite{papineni2002bleu} and TER \cite{Snover:06}}\pp{, computed against the reference translations in the MuST-C En-De test set}.

\section{Results}

We performed preliminary experiments 
\mn{with}
\textit{BASE\_MUSTC} 
\mn{(scoring 21.08 BLEU on the original MuST-C En-De test set)}
to compare the context integration techniques and select the most suitable \mn{one} for ST.
We then compared the fine-tuning  with the context-aware models using the stronger baseline model \textit{BASE\_ALL} (scoring 27.55 BLEU on the original test set).

\begin{table}[htbp]
\caption{\pp{Evaluation results on the VAD-segmented 
test set. Notes: SRC=audio as context; TGT=generated translation as context; SEQ=sequential; PAR=parallel.}}
  \label{tab:results_prelim}
  \centering
  \begin{tabular}{l|c|ccc}
    \toprule
    ~~~ & \textbf{LIUM} & \multicolumn{3}{c}{\textbf{WebRTC}} \\
    ~~~ & ~~~ & \textbf{3, 30ms} & \textbf{2, 20ms} & \textbf{3, 20ms} \\
    \midrule
    BASE\_MUSTC & 17.32 & 17.82 & 17.75 & 16.31\\\hline
    SRC SEQ & 19.08 & 18.81 & 18.00 & 17.42 \\
    SRC PAR & 19.25 & 18.90 & 18.25 & 17.30 \\
    TGT SEQ & 19.57 & \textbf{19.21} & 18.81 & \textbf{17.60} \\
    TGT PAR & \textbf{20.01} & 18.98 & \textbf{18.82} & 17.32 \\
    \bottomrule
  \end{tabular}
  
\end{table}

\begin{table*}[t]
  \caption{Comparison between base model, fine-tuning and context-aware models.}
  \label{tab:results}
  \centering
  \begin{tabular}{l|cc|cccccc}
    \toprule
    ~~~ & \multicolumn{2}{c}{\textbf{LIUM}} & \multicolumn{6}{c}{\textbf{WebRTC}} \\
    ~~~ & ~~~ & ~~~ & \multicolumn{2}{c}{\textbf{AGG=3, FS=30ms}} & \multicolumn{2}{c}{\textbf{AGG=2, FS=20ms}} & \multicolumn{2}{c}{\textbf{AGG=3, FS=20ms}} \\
    \midrule
    ~~~ & BLEU ($\uparrow$) & TER ($\downarrow$) & BLEU ($\uparrow$) & TER ($\downarrow$) & BLEU ($\uparrow$) & TER ($\downarrow$) & BLEU ($\uparrow$) & TER ($\downarrow$) \\
    \midrule
    BASE\_ALL & 19.66 & 76.57 & 22.07 & 67.08 & 21.98 & 66.83 & 19.59 & 72.62 \\
    \midrule
    FINE-TUNE & 22.48 & 64.21 & 23.48 & 60.03 & \textbf{23.40} & 61.54 & 21.35 & 63.90 \\
    \midrule
    TGT SEQ & 23.18 & \textbf{58.60} & 22.85 & \textbf{58.49} & 22.59 & \textbf{59.79} & 21.11 & \textbf{60.51} \\
    \hspace{2mm} + REG & 23.88 & 58.81 & \textbf{23.61} & 58.57 & 23.15 & 60.36 & 21.88 & 60.97 \\
    TGT PAR & 23.77 & 59.02 & 23.34 & 58.94 & 22.91 & 60.09 & 21.75 & 60.77 \\
    \hspace{2mm} + REG & \textbf{23.91} & 58.95 & 23.51 & 58.64 & \textbf{23.40} & 59.95 & \textbf{22.03} & 60.83 \\
    \bottomrule
  \end{tabular}
  
\end{table*}

\subsection{Context information and integration}
\label{sec_res_ctx_type}

\mg{Table  \ref{tab:results_prelim} shows that}
all the tested approaches outperform the baseline
\mg{on VAD-segmented data}
with a margin that ranges from 0.25 to 2.69 BLEU points.
This 
\mg{indicates} that the context is useful to mitigate the effect of  
\mn{VAD-based}
segmentation.
On LIUM, our models 
\pp{achieve}
the highest score (\textit{TGT PAR}, 20.01 BLEU) and \mn{the largest} gain over the baseline;
on WebRTC the improvements are significant
\mn{but smaller.}
We argue that the reason lies in the different characteristics of the two tools.
\mn{The split positions selected by LIUM do not always correspond to actual pauses in the audio, which  prevents the baseline model from disposing of all the information necessary 
for translation.
This information, instead, is available to the context-aware models as they can access the previous segment.}
WebRTC, instead, produces very long/short segments, whose effect on context-aware models is limited: 
 the contribution of adding the previous segment is low both in case of very long segments, as only the first part is influenced by it, and in case of very short ones, as having a short segment as context means adding little information.
We also experimented with including manually-segmented data,
but it was not beneficial for any 
\mn{of our models}.

\mn{Looking at the context modality (text vs audio), we observe that supplying the previously generated translation (TGT*) yields higher BLEU scores than supplying its corresponding audio (SRC*) with both the integration types (*SEQ and *PAR).}
\mn{This suggests that the audio representation produced by current ST models is less suitable than text to extract useful content information to support traslation.
In light of these observations, we decided to proceed with  \textit{TGT SEQ} and \textit{TGT PAR} in the following experiments with the stronger \textit{BASE\_ALL} model.}

\subsection{Context vs fine-tuning}

In this section, we compare the performance of the fine-tuning
and the context-aware solutions. 
In this way, we can disentangle the benefits produced by the context and those due to 
the \mn{use of}
artificial training data. 

The results \mn{in Table \ref{tab:results}} show that: \mn{\textit{i)}}
fine-tuning on the artificial data produces significant gains over \textit{BASE\_ALL} 
\mn{(respectively, 2.82 \mn{BLEU points} on LIUM 
and from 1.41 to 1.76 on WebRTC)}\mn{,
and \textit{ii)}}
\textit{TGT~PAR} outperforms \textit{TGT SEQ} on all datasets (by 0.32 to 0.64).
\textit{TGT PAR} without regularization is superior to the fine-tuning when the VAD splits very aggressively
(21.75 vs 21.35 on WebRTC 3, 20ms) or in non-pause positions (23.77 vs 22.48 on LIUM).
On the other VAD configurations, the results are close, but inferior to the fine-tuning.
Our intuition is that this behavior is caused by the noise added by the context-attention
when the context is not needed. 
This is confirmed by the results obtained adding the context-gate regularization presented in Eq. (\ref{regularization_gate_eq}) (\textit{TGT PAR+REG} and \textit{TGT SEQ+REG}).
The regularization allows our best context-aware model (\textit{TGT PAR+REG}) to outperform the fine-tuned model on 
\pp{3 out of 4}
VAD configurations tested \pp{(in 
\mnn{one}
case 
BLEU is on par)}
and improves both integration types.
\textit{TGT SEQ} benefits more from it, closing the gap with \textit{TGT PAR}.
The value of the hyperparameter $\alpha$ was chosen among $0.01$, $0.02$, $0.04$ and $0.08$:
we set it to $0.04$ as it provided the best loss on the validation set.

The difference between context-aware models and 
fine-tuning is even more evident if we consider the TER \mn{metric (the lower the better)}.
In this case, \textit{TGT SEQ} 
\mn{obtains}
the best scores in 
\mg{every setting},
but the results of all context-aware models are close and are $2$ to $6$ points 
\mn{better}
than those obtained with 
fine-tuning.
We also noticed that {1-},\nolinebreak 2-,3- and 
\mn{4-gram}
BLEU scores are always
\mn{significantly} higher
for the context-aware solutions than for the fine-tuning,
even when the overall BLEU 
\mn{scores are}
similar. The reason lies in the brevity penalty,
as the context-aware models produce shorter translations.
Interestingly, the best result 
(23.91 BLEU) 
is obtained  
\mn{by exploiting}
the context in one of
\mn{the
\mg{hardest} segmentations}
for the base model (19.66 BLEU).
This is coherent with the behavior observed in $\S\ref{sec_res_ctx_type}$.

\section{Analysis}

We performed a manual analysis of the translations produced by the
baseline and by our best context-aware model (\textit{TGT PAR} + REG) on the LIUM-segmented test set. The goal was to check whether the gains are actually due to the use of contextual information and to understand how this information is exploited.
We noticed three main issues solved by 
the context-aware approach.
\mg{They} are all related to the presence of sub-sentential fragments located at the beginning or
the end of a segment.
First,
these fragments are often ignored by the
baseline model. Being trained \pp{only} on well-formed sentences from the clean MuST-C corpus, this model seems unable to handle segments reflecting truncated sentences and, instead of returning partial translations, it  opts for ignoring part of the input audio.
Second, the base model produces \textit{hallucinations} \cite{Lee2018HallucinationsIN}
\mg{trying} to translate a sub-sentential fragment into a well-formed target sentence.
Our models, instead, produce 
the  translation corresponding to the incomplete fragment.
Third, the
baseline model translates the sub-sentential fragment and the adjacent sentence in the same segment into one single output sentence, mixing them.
In contrast, our  models are able to translate them separately.

\section{Conclusions}

\mn{We} studied how to make ST 
\mn{models 
trained on data segmented at sentence-level
robust to
VAD-segmented audio supplied at inference time.}
\mn{To this aim, we 
explored 
different approaches to integrate contextual information 
provided by
the  segment preceding the one to be translated.}
Our experiments show that
adopting
a context-aware architecture,
combined with 
\mn{training on}
artificial data
generated with random segmentation,
\mn{is beneficial to improve final translation quality.}
\pp{We}
also demonstrate
that, compared to the best automatic segmentation (22.07 BLEU), context-aware models
achieve results that are similar in the worst case 
\pp{(22.03)}
and  significantly better in the best case 
\pp{(23.91)}. 
In this case, our context-based approach allows to reduce by 55\% the performance gap of the base model 
\pp{(19.66)}
with respect to  optimal (i.e. sentence-level) manual segmentation 
\pp{(27.55).}
All in all, 
\pp{this suggests}
that working on models' robustness \pp{to sub-optimal VAD segmentation} is at least as promising as improving 
\pp{the segmentation itself.}

\section{Acknowledgements}

This work is part of the ``End-to-end Spoken  Language Translation in Rich Data Conditions'' project,\footnote{https://ict.fbk.eu/units-hlt-mt-e2eslt/} which is financially supported by an Amazon AWS ML Grant.

\bibliographystyle{IEEEtran}

\bibliography{mybib}

\begin{thebibliography}{10}
\providecommand{\url}[1]{#1}
\csname url@samestyle\endcsname
\providecommand{\newblock}{\relax}
\providecommand{\bibinfo}[2]{#2}
\providecommand{\BIBentrySTDinterwordspacing}{\spaceskip=0pt\relax}
\providecommand{\BIBentryALTinterwordstretchfactor}{4}
\providecommand{\BIBentryALTinterwordspacing}{\spaceskip=\fontdimen2\font plus
\BIBentryALTinterwordstretchfactor\fontdimen3\font minus
  \fontdimen4\font\relax}
\providecommand{\BIBforeignlanguage}[2]{{%
\expandafter\ifx\csname l@#1\endcsname\relax
\typeout{** WARNING: IEEEtran.bst: No hyphenation pattern has been}%
\typeout{** loaded for the language `#1'. Using the pattern for}%
\typeout{** the default language instead.}%
\else
\language=\csname l@#1\endcsname
\fi
#2}}
\providecommand{\BIBdecl}{\relax}
\BIBdecl

\bibitem{iwsltprec_2019}
J.~Niehues, R.~Cattoni, S.~Stucker \emph{et~al.}, ``{The IWSLT 2019 Evaluation
  Campaign},'' in \emph{Proc. of 16th International Workshop on Spoken Language
  Translation (IWSLT)}, Hong Kong, Nov. 2019.

\bibitem{berard_2016}
A.~Bérard, O.~Pietquin, C.~Servan, and L.~Besacier, ``{Listen and Translate: A
  Proof of Concept for End-to-End Speech-to-Text Translation},'' in \emph{Proc.
  of NIPS Workshop on end-to-end learning for speech and audio processing},
  Barcelona, Spain, Dec. 2016.

\bibitem{weiss2017sequence}
R.~J. Weiss, J.~Chorowski, N.~Jaitly, Y.~Wu, and Z.~Chen,
  ``{Sequence-to-Sequence Models Can Directly Translate Foreign Speech},'' in
  \emph{Proc. of Interspeech 2017}, Stockholm, Sweden, Aug. 2017, pp.
  2625--2629.

\bibitem{matusov_segm}
E.~Matusov, A.~Mauser, and H.~Ney, ``{Automatic Sentence Segmentation and
  Punctuation Prediction for Spoken Language Translation},'' in \emph{Proc. of
  2006 International Workshop on Spoken Language Translation (IWSLT)}, Kyoto,
  Japan, Nov. 2006, pp. 158--165.

\bibitem{oda-etal-2014-optimizing}
Y.~Oda, G.~Neubig, S.~Sakti, T.~Toda, and S.~Nakamura, ``{Optimizing
  Segmentation Strategies for Simultaneous Speech Translation},'' in
  \emph{Proc. of the 52nd Annual Meeting of the Association for Computational
  Linguistics (ACL)}, Baltimore, Maryland, Jun. 2014, pp. 551--556.

\bibitem{Cho2017}
E.~Cho, J.~Niehues, and A.~Waibel, ``{NMT-Based Segmentation and Punctuation
  Insertion for Real-Time Spoken Language Translation},'' in \emph{Proc. of
  Interspeech 2017}, Stockholm, Sweden, Aug. 2017, pp. 2645--2649.

\bibitem{potapczyk_tomasz_2019}
T.~Potapczyk, P.~Przybysz, M.~Chochowski, and A.~Szumaczuk, ``{Samsung's System
  for the IWSLT 2019 End-to-End Speech Translation Task},'' in \emph{Proc. of
  16th International Workshop on Spoken Language Translation (IWSLT)}, Hong
  Kong, Nov. 2019.

\bibitem{wang-etal-2017-exploiting-cross}
L.~Wang, Z.~Tu, A.~Way, and Q.~Liu, ``{Exploiting Cross-Sentence Context for
  Neural Machine Translation},'' in \emph{Proc. of the 2017 Conference on
  Empirical Methods in Natural Language Processing (EMNLP)}, Copenhagen,
  Denmark, Sep. 2017, pp. 2826--2831.

\bibitem{zhang-etal-2018-improving}
J.~Zhang, H.~Luan, M.~Sun, F.~Zhai, J.~Xu, M.~Zhang, and Y.~Liu, ``{Improving
  the Transformer Translation Model with Document-Level Context},'' in
  \emph{Proc. of the 2018 Conference on Empirical Methods in Natural Language
  Processing (EMNLP)}, Brussels, Belgium, Oct.-Nov. 2018, pp. 533--542.

\bibitem{bawden-etal-2018-evaluating}
R.~Bawden, R.~Sennrich, A.~Birch, and B.~Haddow, ``{Evaluating Discourse
  Phenomena in Neural Machine Translation},'' in \emph{Proc. of the 2018
  Conference of the North {A}merican Chapter of the Association for
  Computational Linguistics: Human Language Technologies (NAACL-HLT)}, New
  Orleans, Louisiana, Jun. 2018, pp. 1304--1313.

\bibitem{kim-etal-2019-document}
Y.~Kim, D.~T. Tran, and H.~Ney, ``{When and Why is Document-level Context
  Useful in Neural Machine Translation?}'' in \emph{Proc. of the 4th Workshop
  on Discourse in Machine Translation (DiscoMT)}, Hong Kong, China, Nov. 2019,
  pp. 24--34.

\bibitem{instance_based}
M.~A. Di~Gangi, V.~Nguyen, M.~Negri, and M.~Turchi, ``{Instance-based Model
  Adaptation for Direct Speech Translation},'' in \emph{Proc. of 2020 IEEE
  International Conference on Acoustics, Speech and Signal Processing
  (ICASSP)}, Barcelona, Spain, May 2020, pp. 7914--7918.

\bibitem{agrawal_ctx}
R.~Agrawal, M.~Turchi, and M.~Negri, ``{Contextual Handling in Neural Machine
  Translation: Look Behind, Ahead and on Both Sides},'' in \emph{Proc. of the
  21st Annual Conference of the European Association for Machine Translation
  (EAMT)}, Alacant, Spain, May 2018, pp. 11--20.

\bibitem{voita-etal-2018-context}
E.~Voita, P.~Serdyukov, R.~Sennrich, and I.~Titov, ``{Context-Aware Neural
  Machine Translation Learns Anaphora Resolution},'' in \emph{Proc. of the 56th
  Annual Meeting of the Association for Computational Linguistics (ACL)},
  Melbourne, Australia, Jul. 2018, pp. 1264--1274.

\bibitem{transformer}
A.~Vaswani, N.~Shazeer, N.~Parmar, J.~Uszkoreit, L.~Jones, A.~N. Gomez,
  L.~Kaiser, and I.~Polosukhin, ``{Attention is All You Need},'' in \emph{Proc.
  of Advances in Neural Information Processing Systems 30 (NIPS)}, Long Beach,
  California, Dec. 2017, pp. 5998--6008.

\bibitem{digangi:interspeech19}
M.~A. Di~Gangi, M.~Negri, and M.~Turchi, ``{Adapting Transformer to End-to-End
  Spoken Language Translation},'' in \emph{Proc. of Interspeech 2019}, Graz,
  Austria, Sep. 2019, pp. 1133--1137.

\bibitem{li2019regularized}
X.~Li, L.~Liu, R.~Wang, G.~Huang, and M.~Meng, ``Regularized context gates on
  transformer for machine translation,'' in \emph{Proc. of the 58th Annual
  Meeting of the Association for Computational Linguistics (ACL)}, Online, Jul.
  2020, pp. 8555--8562.

\bibitem{mustc}
M.~A. Di~Gangi, R.~Cattoni, L.~Bentivogli \emph{et~al.}, ``{MuST-C: a
  Multilingual Speech Translation Corpus},'' in \emph{Proc. of the 2019
  Conference of the North American Chapter of the Association for Computational
  Linguistics: Human Language Technologies (NAACL-HLT)}, Minneapolis,
  Minnesota, Jun. 2019, p. 2012–2017.

\bibitem{dyer-etal-2013-simple}
C.~Dyer, V.~Chahuneau, and N.~A. Smith, ``{A Simple, Fast, and Effective
  Reparameterization of IBM Model 2},'' in \emph{{Proc. of the 2013 Conference
  of the North American Chapter of the Association for Computational
  Linguistics: Human Language Technologies (NAACL-HLT)}}, Atlanta, Georgia,
  Jun. 2013, pp. 644--648.

\bibitem{meigner2010lium}
S.~Meignier and T.~Merlin, ``{LIUM SpkDiarization: An Open Source Toolkit For
  Diarization},'' in \emph{Proc. of the CMU SPUD Workshop}, Dallas, Texas, Mar.
  2010.

\bibitem{szegedy2016rethinking}
C.~Szegedy, V.~Vanhoucke, S.~Ioffe, J.~Shlens, and Z.~Wojna, ``{Rethinking the
  Inception Architecture for Computer Vision},'' in \emph{Proc. of 2016 IEEE
  Conference on Computer Vision and Pattern Recognition (CVPR)}, Las Vegas,
  Nevada, Jun. 2016, pp. 2818--2826.

\bibitem{adam}
D.~Kingma and J.~Ba, ``{Adam: A Method for Stochastic Optimization},'' in
  \emph{Proc. of 3rd International Conference on Learning Representations
  (ICLR)}, San Diego, California, May 2015.

\bibitem{Park_2019}
D.~S. Park, W.~Chan, Y.~Zhang, C.-C. Chiu, B.~Zoph, E.~D. Cubuk, and Q.~V. Le,
  ``{SpecAugment: A Simple Data Augmentation Method for Automatic Speech
  Recognition},'' in \emph{Proc. of Interspeech 2019}, Graz, Austria, Sep.
  2019, pp. 2613--2617.

\bibitem{gaido-etal-2020-end}
M.~Gaido, M.~A. Di~Gangi, M.~Negri, and M.~Turchi, ``{End-to-End
  Speech-Translation with Knowledge Distillation: {FBK}@{IWSLT}2020},'' in
  \emph{Proc. of the 17th International Conference on Spoken Language
  Translation (IWSLT)}, Online, Jul. 2020, pp. 80--88.

\bibitem{ott-etal-2019-fairseq}
M.~Ott, S.~Edunov, A.~Baevski, A.~Fan, S.~Gross, N.~Ng, D.~Grangier, and
  M.~Auli, ``{fairseq: A Fast, Extensible Toolkit for Sequence Modeling},'' in
  \emph{Proc. of the 2019 Conference of the North American Chapter of the
  Association for Computational Linguistics: Human Language Technologies
  (NAACL-HLT)}, Minneapolis, Minnesota, Jun. 2019, pp. 48--53.

\bibitem{koehn-etal-2007-moses}
P.~Koehn, H.~Hoang, A.~Birch \emph{et~al.}, ``{Moses: Open Source Toolkit for
  Statistical Machine Translation},'' in \emph{Proc. of the 45th Annual Meeting
  of the Association for Computational Linguistics (ACL)}, Prague, Czech
  Republic, Jun. 2007, pp. 177--180.

\bibitem{sennrich2015neural}
R.~Sennrich, B.~Haddow, and A.~Birch, ``{Neural Machine Translation of Rare
  Words with Subword Units},'' in \emph{Proc. of the 54th Annual Meeting of the
  Association for Computational Linguistics (ACL)}, Berlin, Germany, Aug. 2016,
  pp. 1715--1725.

\bibitem{neubig-etal-2018-xnmt}
G.~Neubig, M.~Sperber, X.~Wang \emph{et~al.}, ``{{XNMT}: The e{X}tensible
  Neural Machine Translation Toolkit},'' in \emph{Proc. of the 13th Conference
  of the Association for Machine Translation in the {A}mericas (AMTA)}, Boston,
  Massachusetts, Mar. 2018, pp. 185--192.

\bibitem{papineni2002bleu}
K.~Papineni, S.~Roukos, T.~Ward, and W.-J. Zhu, ``{BLEU: a Method for Automatic
  Evaluation of Machine Translation},'' in \emph{Proc. of the 40th Annual
  Meeting of the Association for Computational Linguistics (ACL)},
  Philadelphia, Pennsylvania, Jul. 2002, pp. 311--318.

\bibitem{Snover:06}
M.~Snover, B.~Dorr, R.~Schwartz, L.~Micciulla, and J.~Makhoul, ``{A Study of
  Translation Edit Rate with Targeted Human Annotation},'' in \emph{Proc. of
  the 7th Conference of the Association for Machine Translation in the Americas
  (AMTA)}, Cambridge, Massachusetts, Aug. 2006, pp. 223--231.

\bibitem{Lee2018HallucinationsIN}
K.~Lee, O.~Firat, A.~Agarwal, C.~Fannjiang, and D.~Sussillo, ``{Hallucinations
  in Neural Machine Translation},'' in \emph{Proc. of NIPS 2018
  Interpretability and Robustness for Audio, Speech and Language Workshop},
  Montr{\'e}al, Canada, Dec. 2018.

\end{thebibliography}


\end{document}